\documentclass[journal]{IEEEtran}
\usepackage{float}
\usepackage{hyperref}
\hypersetup{
    colorlinks=true,
    linkcolor=blue,    
    citecolor=blue,      
    urlcolor=blue       
}

\usepackage[
backend=biber,
style=numeric,
sorting=none
]{biblatex}
\usepackage{amsmath}
\usepackage{amsfonts}
\usepackage{algorithm}
\usepackage{tabularx}
\usepackage{algorithmicx}
\usepackage{booktabs}   
\usepackage{multirow}
\usepackage[caption=false,font=footnotesize]{subfig}
\usepackage{graphicx} 
\usepackage{algpseudocode}
\usepackage{subcaption}
\usepackage{xcolor}
\usepackage{tikz}
\usepackage[caption=false,font=footnotesize]{subfig}
\usetikzlibrary{arrows.meta,shapes.geometric,shapes.misc}

\definecolor{OISky}{HTML}{56B4E9}
\definecolor{OIGreen}{HTML}{009E73}
\definecolor{OIBlue}{HTML}{004B97}
\definecolor{OIVerm}{HTML}{D55E00}
\definecolor{OIPurple}{HTML}{CC79A7}
\definecolor{OIYellow}{HTML}{F0E442}
\definecolor{cClean}{HTML}{00b050}    
\definecolor{cAdv}{HTML}{ff0000}      
\definecolor{cEnc}{HTML}{a5efff}      
\definecolor{cLat}{HTML}{628cff}      
\definecolor{cDec}{HTML}{66d8c7}      
\definecolor{cConvGN}{HTML}{ff7e79}   
\definecolor{cTE}{HTML}{d36efa}       
\definecolor{cSkip}{HTML}{9b9b9b}     
\definecolor{cDown}{HTML}{b0b4df}     
\definecolor{cUp}{HTML}{6c42f3}       
\definecolor{cAlignAdd}{HTML}{e1b348} 
\definecolor{cHead}{HTML}{9b2d61}     

\DeclareRobustCommand{\legSquare}[2]{%
  \begin{tikzpicture}[baseline=-0.6ex]
    \node[draw, rectangle, fill=#1, minimum width=4mm, minimum height=4mm] {};
  \end{tikzpicture}~#2%
}
\DeclareRobustCommand{\legRound}[2]{%
  \begin{tikzpicture}[baseline=-0.6ex]
    \node[draw, rounded corners=2pt, fill=#1, minimum width=4mm, minimum height=4mm] {};
  \end{tikzpicture}~#2%
}
\DeclareRobustCommand{\legArrow}[2]{%
  \begin{tikzpicture}[baseline=-0.6ex]
    \draw[->, line width=0.9pt, #1] (0,0)--(8mm,0);
  \end{tikzpicture}~#2%
}
\DeclareRobustCommand{\legDash}[1]{%
  \begin{tikzpicture}[baseline=-0.6ex]
    \draw[dashed, line width=0.9pt] (0,0)--(8mm,0);
  \end{tikzpicture}~#1%
}
\DeclareRobustCommand{\legSync}[1]{%
  \begin{tikzpicture}[baseline=-0.6ex]
    \draw[<->, dashed, line width=0.9pt] (0,0)--(8mm,0);
  \end{tikzpicture}~#1%
}
\DeclareRobustCommand{\legDiamond}[2][]{%
  \begin{tikzpicture}[baseline=-0.6ex]
    \node[draw, diamond, aspect=1.0, minimum width=4.4mm, minimum height=3.2mm,
          fill=OIYellow!45,#1] {};
  \end{tikzpicture}~#2%
}

\begin{document}

\title{MedFedPure: A Medical Federated Framework with MAE-based Detection and Diffusion Purification for Inference-Time Attacks}

\author{
Mohammad Karami\textsuperscript{1},
Mohammad Reza Nemati\textsuperscript{1,$\dagger$},
Aidin Kazemi\textsuperscript{1,$\dagger$},
Ali Mikaeili Barzili\textsuperscript{1,2},
Hamid Azadegan\textsuperscript{3},
and Behzad Moshiri\textsuperscript{1,4,*} \\
\IEEEauthorblockA{\textsuperscript{1}\textit{School of Electrical and Computer Engineering}, University of Tehran, Tehran, Iran \\
\{mohammad.karami79, mohammad.r.nemati, aidin.kazemi, mikaeili.ali, moshiri\}@ut.ac.ir}\\
\IEEEauthorblockA{\textsuperscript{2}\textit{Max Planck Institute for Brain Research}, Frankfurt, Germany \\
Ali.mikaeili@brain.mpg.de} \\
\IEEEauthorblockA{\textsuperscript{3}\textit{School of Computer Engineering,University of Science and Technology (IUST)}, Tehran, Iran \\
H\_azadegan@vu.iust.ac.ir} \\
\IEEEauthorblockA{\textsuperscript{4}\textit{Department of Electrical and Computer Engineering}, University of Waterloo, Ontario, Canada \\
bmoshiri@uwaterloo.ca}
\thanks{\textsuperscript{*}Corresponding author}
\thanks{\textsuperscript{$\dagger$}These authors contributed equally to this work.}
}

\maketitle
\begin{abstract}
Artificial intelligence (AI) has shown great potential in medical imaging, particularly for brain tumor detection using Magnetic Resonance Imaging (MRI). However, the models remain vulnerable at inference time when they trained collaboratively through Federated Learning (FL), an approach adopted to protect patient privacy. Adversarial attacks can subtly alter medical scans in ways invisible to the human eye yet powerful enough to mislead AI models, potentially causing serious misdiagnoses. Existing defenses often assume centralized data and struggle to cope with the decentralized and diverse nature of federated medical settings.
In this work, we present MedFedPure, a personalized federated learning defense framework, designed to protect diagnostic AI models at inference time without compromising privacy or accuracy. MedFedPure combines three key elements: (1) a personalized FL model that adapts to the unique data distribution of each institution; (2) a Masked Autoencoder (MAE) that detects suspicious inputs by exposing hidden perturbations; and (3) an adaptive, diffusion-based purification module that selectively cleans only the flagged scans before classification. Together, these steps offer robust protection while preserving the integrity of normal, benign images.
We evaluated MedFedPure on the Br35H brain MRI dataset. The results show a significant gain in adversarial robustness, improving performance from 49.50\% to 87.33\% under strong attacks, while maintaining a high clean accuracy of 97.67\%. By operating locally and in real time during diagnosis, our framework provides a practical path to deploying secure, trustworthy, and privacy-preserving AI tools in clinical workflows.
\end{abstract}

\IEEEpeerreviewmaketitle

\section{Introduction}

Cancer remains one of the most pressing global health challenges, claiming millions of lives each year~\cite{who_cancer}. Among these, brain and central nervous system tumors, though relatively less common, carry a disproportionately high risk of morbidity and mortality due to their direct impact on neurological function. Early and accurate diagnosis is therefore critical to guide therapy and improve patient outcomes. Magnetic resonance imaging (MRI) has become the clinical standard for non-invasive brain tumor detection because of its high-resolution imaging and superior soft-tissue contrast.

Yet, the interpretation of MRI scans is time-consuming and prone to inter-observer variability, creating an urgent need for reliable computer-aided diagnostic (CAD) systems. Deep learning models, particularly convolutional networks and encoder-decoder architectures such as ResNet and U-Net variants, have shown remarkable improvements in tumor detection and segmentation, bringing consistency and efficiency to radiological workflows. However, these models depend on access to large, diverse, and well-annotated datasets that are challenging to centralize in medical domains due to strict privacy regulations and institutional barriers.

\begin{figure}[!t]
    \centering
    \includegraphics[width=\columnwidth]{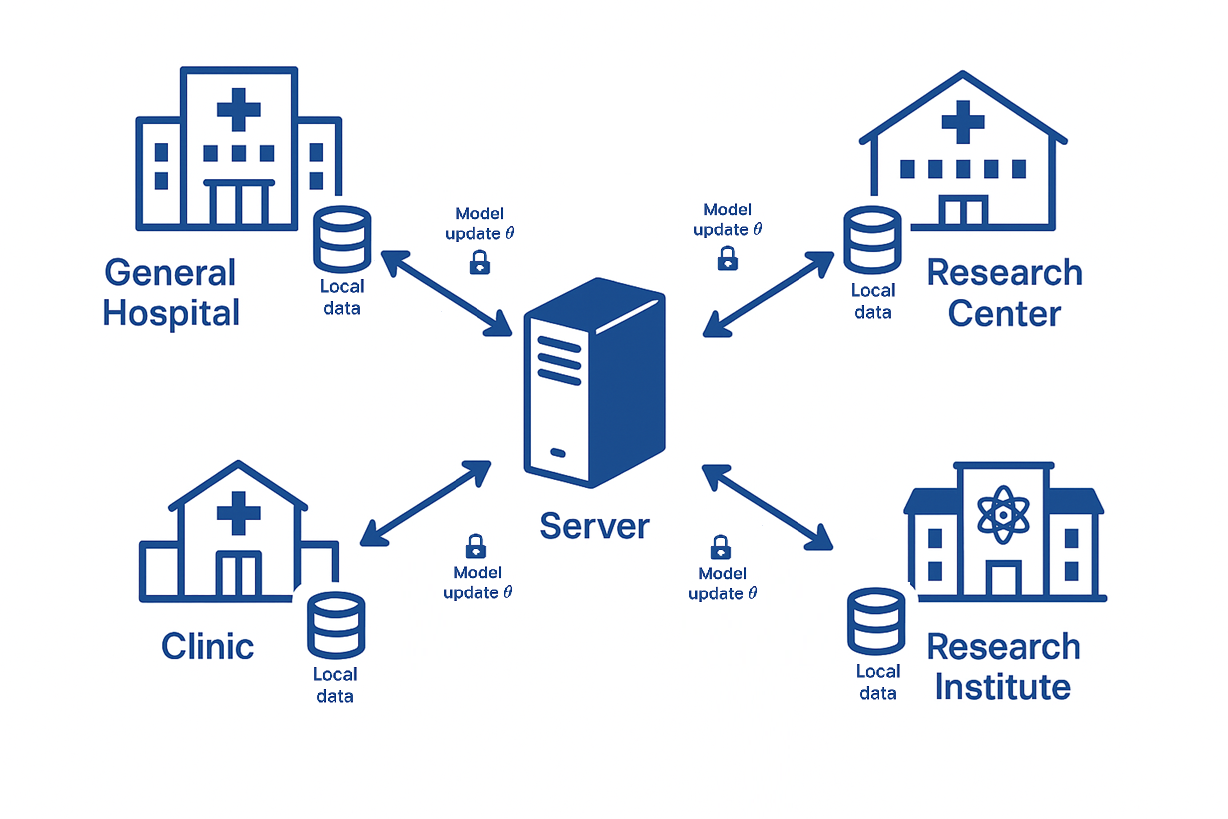} 
    \caption{Federated learning overview in our setting: each site trains locally and keeps \emph{local data} on-premise; only model updates $\theta$ are sent to the server for aggregation (e.g., FedAvg), and the updated model is returned to clients.}
    \label{fig:fl_overview}
\end{figure}

Federated Learning (FL) has emerged as a compelling solution to this data challenge (see Fig.~\ref{fig:fl_overview}). By enabling institutions to collaboratively train models without sharing raw patient data, FL strikes a balance between building robust models and preserving privacy. Successful demonstrations, such as the FeTS initiative, show that federated models can approach the accuracy of centrally trained ones while respecting institutional boundaries. This positions FL as a promising paradigm for cross-institutional collaboration in healthcare.

Despite this promise, two unresolved issues remain. First, the statistical heterogeneity of data across clients often degrades global model performance. Personalized Federated Learning (PFL) has been introduced to address this by tailoring models to individual data distributions while maintaining collaborative benefits. Second, and more critically for clinical safety, FL models are vulnerable at inference time. In the context of federated learning, the notion of an \textit{attacker} refers to a malicious entity that manipulates model inputs at inference. Such \textit{attacks} do not require access to patient records or the federated training process; instead, they operate by crafting subtle perturbations to medical scans that remain imperceptible to clinicians but mislead the model. For instance, an attacker could modify a brain MRI so that a tumor-bearing image is classified as normal, or conversely, cause a healthy scan to be flagged as malignant. Most existing defenses have been designed for centralized training settings or focus on training-time robustness, incurring heavy computational costs and leaving inference-time security largely unaddressed. Moreover, personalization complicates the defense landscape: a uniform protection mechanism may unfairly affect clients with unique data characteristics.

This work addresses this unmet need by proposing a federated, inference-time defense framework that is adaptive, efficient, and client-specific. Our framework operates exclusively on the client side, protecting local inference without altering federated training or sharing sensitive information. It integrates three components: (i) a personalized mixture-of-experts classifier that adapts to each client's data distribution, (ii) a Masked Autoencoder (MAE) detector that identifies potentially adversarial inputs through reconstruction error, and (iii) a diffusion-based purification module that selectively sanitizes flagged samples, with purification strength adjusted according to the detector's confidence. This adaptive strategy ensures that benign inputs are preserved, while adversarial perturbations are effectively neutralized.

The main contributions of this paper are:
\begin{enumerate}
    \item We introduce a novel client-side defense framework that combines personalized federated learning with a multi-stage, test-time protection pipeline against adversarial attacks. 
    \item We design an adaptive detection-purification mechanism, where an MAE-based detector triggers conditional diffusion-based purification, balancing security with efficiency and fidelity on clean inputs.
    \item We provide comprehensive empirical validation on standard benchmark (CIFAR-10) and a clinically relevant MRI dataset (BR35H), demonstrating that our approach significantly enhances adversarial robustness while maintaining high accuracy on benign data.
\end{enumerate}
The remainder of the paper is organized as follows. Section~\ref{sec:related} reviews related work on federated personalization and test-time defenses. Section~\ref{sec:method} details the proposed personalized federated learning architecture, the MAE detector, and the diffusion-based purifier. Section~\ref{sec:experiments} describes experimental setup and evaluation results, and Section~\ref{sec:conclusion} concludes with discussion and future directions.

\section{Literature Review}\label{sec:related}

\subsection{Brain Tumor Detection and MRI Imaging}

Malignancies of the brain and central nervous system, although less common than other cancers, account for nearly 2\% of all diagnoses and contribute to over 2.5\% of cancer-related mortality worldwide \cite{Ilic2023}. Brain tumors, which manifest as abnormal cell proliferation within the cranial cavity, exert pressure on surrounding neural structures and can significantly impair neurological function \cite{Raghavendra2023-ky, ABDELLAH2019300}. 

The potential for rapid tumor progression requires early detection to maximize treatment efficacy \cite{s24185953, Raghavendra2023-ky}. Among diagnostic modalities, Magnetic Resonance Imaging (MRI) has established itself as a non-invasive technique offering high-resolution imaging of brain structures, making it invaluable for tumor localization and classification \cite{Firmansyah_Utama_2024}.

Manual interpretation of MRI scans is both labor-intensive and subject to inter-observer variability. To address this, computer-aided diagnostic (CAD) tools have gained prominence, aiming to automate and enhance the processes of tumor identification, segmentation, and classification \cite{ABDELLAH2019300}. Recent advances have focused on deep learning architectures, including variants of ResNet and U-Net, which have demonstrated significant improvements in accuracy, computational efficiency, and clinical applicability \cite{Aggarwal2023, Munir2021, Munir2021_2, AbdElKader2021, 10880261}.
The development of reliable AI models requires access to curated and representative datasets. Among the publicly available brain MRI collections, the Br35H Brain Tumor Detection dataset has been frequently utilized due to its well-labeled samples and practical relevance to real-world clinical tasks \cite{tbkk-q937-25}. Several studies have benchmarked its use in training deep learning frameworks for tumor identification, demonstrating its efficacy in enhancing diagnostic accuracy \cite{https://doi.org/10.1155/2021/5513500, chhatre2023brain, khan2024comparativeanalysisresourceefficientcnn, 10286729}.

\subsection{Federated Learning in Medical Imaging}

Federated Learning (FL) addresses the critical privacy and data governance challenges inherent in multi-institutional medical data analysis \cite{Dayan2021, LI2020106854}. In this framework, multiple clients independently train a shared model using local data and periodically synchronize with a central server via model updates rather than raw data. The widely adopted FedAvg algorithm \cite{pmlr-v54-mcmahan17a} facilitates global model aggregation through weighted averaging of local parameters.

FL can be classified based on data distribution into horizontal (HFL), vertical (VFL), and federated transfer learning (FTL). HFL assumes shared feature spaces with disjoint user sets, while VFL is suitable when clients share samples but differ in features. FTL accommodates heterogeneity in both samples and features, leveraging transfer learning to bridge representational gaps \cite{10274102, 10.1145/3298981, jahani2024survey}.

Over time, FL has gained traction as a privacy-preserving ML framework, driving a surge of theoretical studies, real-world deployments, and clinical interest \cite{NIPS2017_6211080f, a9492000}. Both industrial and academic sectors have pursued its application in sensitive domains such as medical imaging, particularly for decentralized analysis of brain tumor data \cite{Zhou_Wang, Albalawi2024}. The sensitive nature of medical data has positioned FL as a valuable solution for cross-institutional collaboration without compromising patient confidentiality \cite{Rieke2020}.
Applications in neuroimaging have highlighted the promise of FL in constructing robust models across decentralized medical repositories \cite{pfitzner2021federated}. A notable example is the FeTS Initiative \cite{Pati2022}, which showed that FL could match the segmentation performance of centralized models across multiple clinical institutions, without ever sharing raw MRI data. 

\subsection{Challenges and Personalization in Federated Learning}

Despite its advantages, FL faces several key limitations. The vulnerability to inference-time adversarial attacks, where malicious queries are crafted to mislead the model or extract sensitive knowledge, poses security threats \cite{barreno2006can, jahani2024ppfl, 9945997}. These attacks vary based on the attacker's knowledge, from full-access white-box scenarios to black-box settings where only model outputs are observable \cite{REN2020346, jahani2025secure}.

Additionally, the inherent data heterogeneity in FL systems contributes to degraded performance, especially when client data distributions differ significantly from the global average. The noise introduced through privacy-preserving techniques such as differential privacy further exacerbates this trade-off between model accuracy and confidentiality, often resulting in bias or fairness concerns \cite{pmlr-v84-bellet18a, GU2022102907, karami2025optigradtrust}.

To mitigate these drawbacks, the field of Personalized Federated Learning (PFL) has emerged. PFL tailors models to the individual characteristics of each client while maintaining collaborative training benefits \cite{10.1016/j.eswa.2023.122874}. Approaches to personalization include local fine-tuning, multi-task learning \cite{arivazhagan2019federatedlearningpersonalizationlayers}, and meta-learning frameworks \cite{fallah2020personalizedfederatedlearningmetalearning}. Solutions such as FedBN \cite{li2021fedbn} 
and SCAFFOLD \cite{pmlr-v119-karimireddy20a} address client drift and non-IID issues by decoupling shared representations from client-specific layers or by correcting local update bias. 

Personalization strategies are broadly categorized into three types: model-based (e.g., personalized heads or layers) 
\cite{Ma_2022_CVPR}, data-based (e.g., client clustering) \cite{mansour2020approachespersonalizationapplicationsfederated, NEURIPS2020_e32cc80b}, and optimization-based techniques (e.g., regularization or meta-learning) \cite{Tan_Long_LIU_Zhou_Lu_Jiang_Zhang_2022}. These strategies collectively aim to improve convergence, fairness, and user experience in federated systems.





\subsection{Inference-Time Robustness and Security}

Even after model deployment, inference-time vulnerabilities remain an open concern. Test-time attacks can significantly compromise the integrity of the model \cite{Wu2024Survey, Ye2024BapFL}. To protect deployed models, researchers have proposed a range of post-training defenses. These methods operate on the input of the model or on its output. Input-level strategies aim to detect or neutralize malicious inputs before they affect the model. For instance, STRIP perturbs incoming samples with noise and checks the prediction consistency to expose Trojan triggers \cite{gao2019strip}.
Similarly, the TeCo approach applies input corruptions and monitors the robustness of the model's predictions, flagging inputs that cause abnormal output changes \cite{Liu2023TeCo}. Beyond detection, generative refinement techniques have been explored to sanitize inputs: diffusion-based purification methods leverage powerful generative models to remove adversarial perturbations from images without requiring any model modifications, achieving state-of-the-art resilience against a variety of attacks \cite{Nie2022DiffPure}. 

In contrast, output-level defenses focus on identifying and mitigating attacks by analyzing the model's prediction patterns. 
The NAB technique introduces a benign decoy trigger into the model during deployment, which helps nullify or overwrite any malicious backdoor activation at inference \cite{Liu2023NAB}. Another line of work integrates external generative models into the prediction pipeline: ZIP uses a pre-trained diffusion model to transform a suspect input (e.g., via blurring and subsequent generative reconstruction) such that any embedded trigger is erased, thereby restoring the model's correct output without requiring prior knowledge of the attack \cite{Shi2023ZIP}. These diverse approaches illustrate the growing toolkit for test-time defense in deep learning. Such mechanisms are especially crucial in medical AI applications, where an undetected adversarial or backdoor attack on a diagnostic model could lead to a dangerously incorrect decision.

\subsection{Research Gap Analysis}

Despite progress in federated learning and in adversarial defense, significant gaps remain at their intersection. Notably, most inference-time defense techniques have been developed and evaluated in centralized settings, assuming a static model with access to all data. In federated environments, however, data and model heterogeneity pose unique challenges that existing defenses do not fully address. 
Effective countermeasures for inference-stage attacks in FL are largely lacking. Current federated defense research tends to focus on training-time robustness (e.g., robust aggregation or adversarial training), which can incur high communication and computation costs and may degrade model accuracy. In contrast, a client-side, test-time defense offers the promise of enhancing security on demand without altering the federated training process or sharing sensitive data \cite{10983777}.

Moreover, personalization in FL complicates the defense landscape: a one-size-fits-all adversarial filter could unfairly impact certain clients due to their unique data distributions. This calls for defense strategies that can adapt to individual clients. An approach by Tsai et al. \cite{tsai2023testtimedetectionrepairadversarial} introduced a masked autoencoder-based technique to detect and repair adversarial samples at test time on a fixed, centrally trained model. However, no prior work (to our knowledge) has integrated such test-time detection and purification into a federated, personalized learning context. In summary, there is a need for a federated inference-time defense framework that preserves privacy, accommodates client-specific model variations, and provides robust protection against adversarial and backdoor threats. This gap motivates our proposed method, which combines personalized FL with on-device adversarial detection and input purification to secure brain tumor classification in a distributed setting.

\section{Methodology}\label{sec:method}

\begin{figure*}[!t]
\centering

\subfloat[Overall system.]{%
  \includegraphics[width=.49\linewidth]{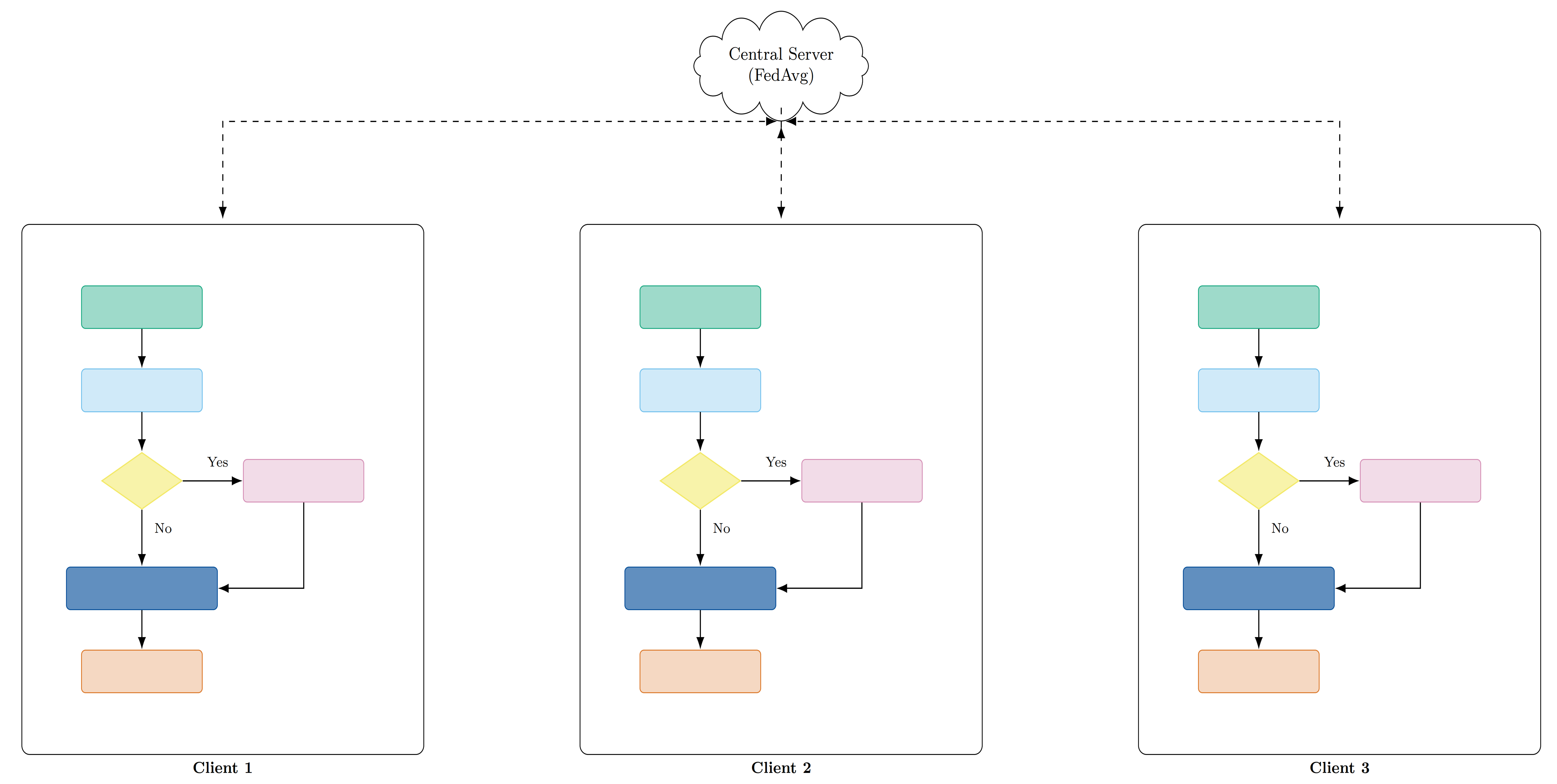}%
  \label{fig:overall_sys}}
\hfill
\subfloat[Client view.]{%
  \includegraphics[width=.49\linewidth]{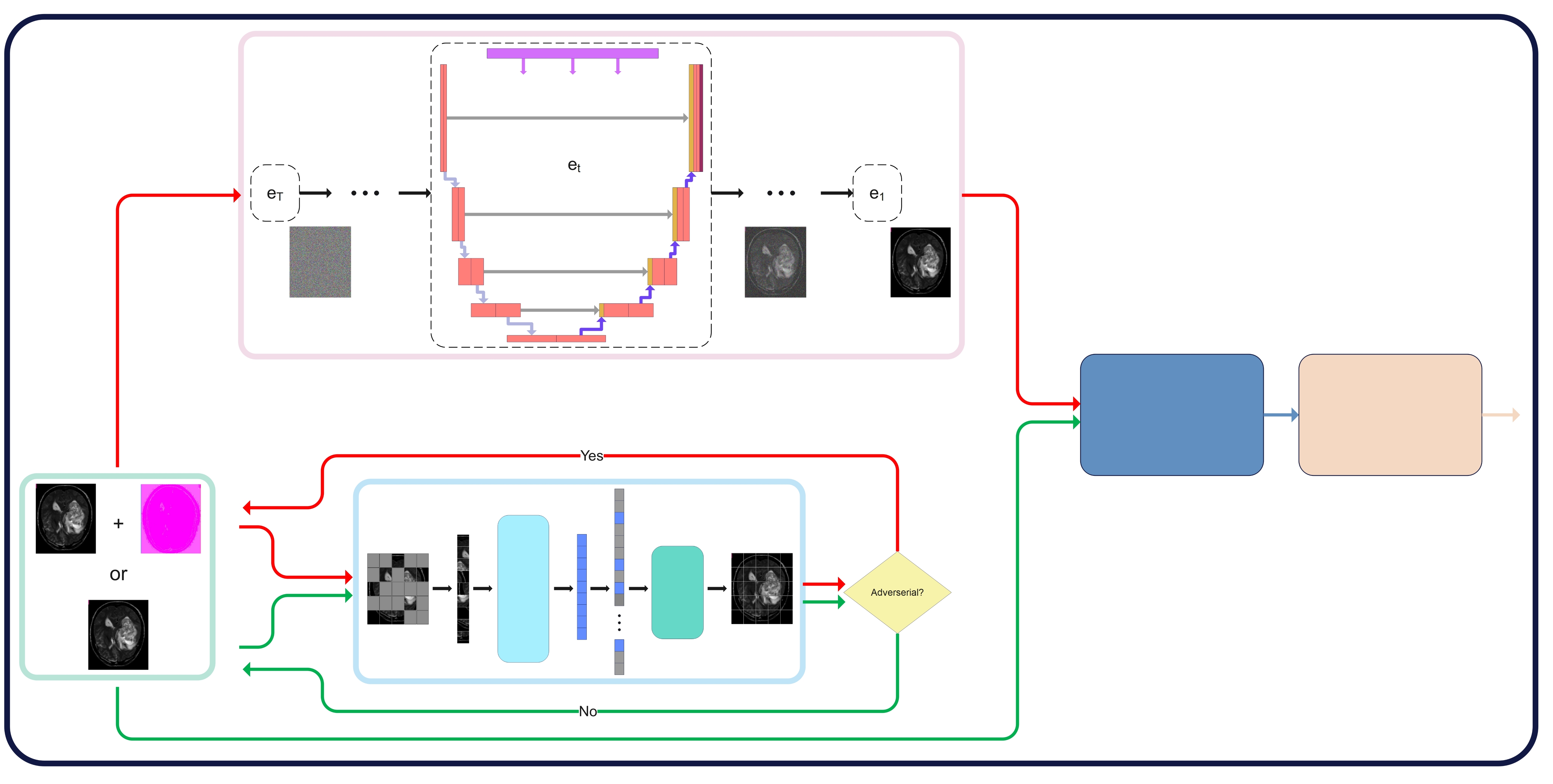}%
  \label{fig:client_view}}

\vspace{4mm}

\footnotesize
\begin{minipage}{\linewidth}
\setlength{\tabcolsep}{6pt}
\renewcommand{\arraystretch}{1.08}
\begin{tabular}{@{}p{0.30\linewidth} p{0.18\linewidth} p{0.18\linewidth} p{0.24\linewidth}@{}}
\legRound{cEnc}{MAE encoder} &
\legSquare{cLat}{Latent tokens} &
\legRound{cDec}{MAE decoder} &
\legSquare{cTE}{Time embedding $e_t$ (at GN)} \\
\legSquare{cConvGN}{Conv $3{\times}3 \to$ GN $(\text{from } e_t) \to$ SiLU} &
\legSquare{cAlignAdd}{1$\times$1 align $\oplus$ add} &
\legSquare{cHead}{1$\times$1 $\rightarrow \varepsilon_{\phi}(x_t,t)$} &
\legRound{OIGreen!38}{Input $x$} \\
\legRound{OISky!28}{Detector $s_{\mathrm{det}}(x)$} &
\legDiamond{Decision $(>\tau)$} &
\legRound{OIPurple!26}{Purifier} &
\legRound{OIBlue!62}{Classifier} \\
\legRound{OIVerm!24}{Output $\hat{y}$} &
\legArrow{cClean}{Clean path} &
\legArrow{cAdv}{Adversarial path} &
\legArrow{cSkip}{Skip} \\
\legArrow{cDown}{Down $\times 2$} &
\legArrow{cUp}{Up $\times 2$} &
\legSync{Federated sync} &
\legDash{Diffusion stages} \\
\end{tabular}
\end{minipage}

\caption{Overview of the proposed federated defense framework. (a) Overall architecture with FedAvg; (b) client workflow integrating MAE detection, diffusion purification (if $s_{\mathrm{det}}(x)>\tau$), and classification.}
\label{fig:overall-client-shared-legend}
\end{figure*}

The proposed methodology introduces a layered adversarial defense framework for federated learning, designed to operate exclusively at test-time on the client side (see Fig.~\ref{fig:overall-client-shared-legend}a). The approach is structured as a three-stage pipeline that combines a personalized FL backbone with specialized detection and purification modules.

At its core, the classification component is derived from the Personalized Federated Defense (\textit{pFedDef}) \cite{kim2023characterizing} paradigm. In this setting, each client maintains a mixture-of-experts model comprising $K = 3$ learners, each instantiated as a ResNet-18 \cite{he2016deep} architecture. The outputs of these learners are modulated by an input-dependent attention mechanism, which enables each client to adapt its inference to the unique characteristics of its local data distribution. Model updates are aggregated at the server using the FedAvg algorithm, ensuring both personalization and global knowledge sharing.

The defense sequence (Fig.~\ref{fig:overall-client-shared-legend}b) is activated whenever a client receives a test input. Initially, an adversarial detection module, implemented as a Masked Autoencoder (MAE) \cite{he2022masked} with a Vision Transformer (ViT) \cite{dosovitskiy2020image} backbone, estimates a reconstruction error for the given input. Inputs whose error exceeds a predefined threshold $\tau$ are flagged as potentially adversarial. These flagged inputs are subsequently processed by a diffusion-based purification module constructed upon a U-Net \cite{ronneberger2015u} architecture. This module executes a partial reverse diffusion process, effectively attenuating adversarial perturbations. The number of purification timesteps is adaptively determined according to the detection confidence score generated by the MAE, yielding a computationally efficient defense mechanism. 

Following detection and, when necessary, purification, the resulting input, either original or sanitized, is forwarded to the client's model for final classification. This integrated design provides resilience against $\ell_\infty$-bounded adversarial attacks, such as Projected Gradient Descent (PGD) \cite{madry2018towards}, by first identifying high-risk samples and then selectively mitigating them.

\subsection{Problem Formulation}

\subsubsection{Federated Learning Setup}

We consider a FL environment consisting of $N$ clients, where each client $i$ maintains a private dataset $\mathcal{D}_i = \{(x_j^{(i)}, y_j^{(i)})\}_{j=1}^{n_i}$ containing $n_i$ labeled samples. The aim of the system is to collaboratively train a global model without directly sharing raw data, thereby preserving data privacy.

The overall optimization problem can be expressed as:
\begin{equation}
    \min_{\theta} F(\theta) = \sum_{i=1}^{N} \frac{n_i}{n} F_i(\theta),
\end{equation}
where $F_i(\theta) = \frac{1}{n_i} \sum_{j=1}^{n_i} \ell(f_{\theta}(x_j^{(i)}), y_j^{(i)})$ denotes the local empirical risk on client $i$, $n = \sum_{i=1}^{N} n_i$ is the total number of samples across all clients, $\ell(\cdot, \cdot)$ represents the cross-entropy loss function, and $f_\theta(\cdot)$ is the model parameterized by $\theta$.

\subsubsection{Adversarial Threat Model}

In this work, we assume the presence of adversaries capable of crafting perturbations constrained by an $\ell_\infty$ norm bound. An adversarial example for a benign input $x$ can be defined as:
\begin{equation}
    x_{\mathrm{adv}} = x + \delta, \quad \text{s.t.} \quad \|\delta\|_\infty \leq \epsilon,
\end{equation}
where $\delta$ is the perturbation vector and $\epsilon$ controls the maximum perturbation magnitude.

We adopt the widely-used PGD method to model the adversary's strategy. The iterative generation process is given by:
\begin{equation}
    x_{\mathrm{adv}}^{(t+1)} = \Pi_{\mathcal{B}_\epsilon(x)} \left( x_{\mathrm{adv}}^{(t)} + \alpha \cdot \mathrm{sign} \big( \nabla_x \ell(f_\theta(x_{\mathrm{adv}}^{(t)}), y) \big) \right),
\end{equation}
where $\Pi_{\mathcal{B}_\epsilon(x)}$ denotes projection onto the $\ell_\infty$ ball of radius $\epsilon$ centered at $x$, $\alpha$ is the step size, and $t$ indexes the attack iterations.

This formulation captures a worst-case scenario in which the adversary has full knowledge of the model parameters and can exploit gradient information to maximize prediction errors.

\subsection{Proposed Defense Framework}

This section introduces our \textit{Personalized Federated Defense} method, a hybrid defense framework combining personalized multi-expert learning, diffusion-based input purification, and masked autoencoder-based adversarial detection. The overall design aims to simultaneously enhance robustness against adversarial perturbations and maintain high performance on benign data in FL environments.

\subsubsection{Personalized Federated Defense}

The core idea of our PFL method is a mixture-of-experts formulation, where each client maintains $K = 3$ local learners with dynamic attention mechanisms:
\begin{equation}
    \theta_i = \{\theta_i^{(1)}, \theta_i^{(2)}, \theta_i^{(3)}\}, \quad \phi_i = \{\phi_i^{(1)}, \phi_i^{(2)}, \phi_i^{(3)}\}
\end{equation}
where $\theta_i^{(k)}$ represents the parameters of the $k$-th learner and $\phi_i^{(k)}$ represents the attention network parameters.
The local prediction is obtained via a weighted combination of learner outputs:
\begin{equation}
    \hat{y}_i(x) = \sum_{k=1}^{K} \alpha_i^{(k)}(x) \cdot f_{\theta_i^{(k)}}(x),
\end{equation}

where the attention weights $\alpha_i$ are computed through a softmax-normalized attention mechanism:

\begin{equation}
\alpha_i^{(k)}(x) = \frac{\exp(g_{\phi_i^{(k)}}(\psi(x)))}{\sum_{j=1}^{K} \exp(g_{\phi_i^{(j)}}(\psi(x)))}.
\end{equation}
Here, $\psi(x) \in \mathbb{R}^{d}$ represents the shared feature representation extracted by the ResNet-18 backbone, and $g_{\phi_i^{(k)}}$ is a learner-specific attention network.

The parameters of the expert networks and the attention networks for each client i are jointly optimized by minimizing the cross-entropy loss on the final aggregated prediction:
\begin{equation}
\mathcal{L}_{\text{pre-reg}} = \ell(\hat{y}_i(x), y) = \ell\left(\sum{k=1}^{K} \alpha_i^{(k)}(x) \cdot f{\theta_i^{(k)}}(x), y\right).
\label{eq:client_loss}
\end{equation}
This objective encourages the attention mechanism to assign higher weights to the expert best suited for a given input, thereby specializing each learner while contributing to a robust ensemble.

Each local learner comprises:
\begin{itemize}
    \item \textbf{Feature Extractor:} ResNet-18 backbone with ImageNet-pretrained weights.
    \item \textbf{Classifier Head:} Two-layer MLP with 256 hidden units and ReLU activation.
    \item \textbf{Attention Network:} Two-layer MLP producing scalar attention scores.
\end{itemize}

The mixture weights are obtained from a softmax over attention network outputs:
\begin{equation}
    w_i^{(k)}(x) = \frac{\exp(\mathrm{MLP}_i^{(k)}(\psi(x)))}{\sum_{j=1}^{K} \exp(\mathrm{MLP}_i^{(j)}(\psi(x)))}.
\end{equation}
The training objective for client $i$ combines classification loss with attention regularization:
\begin{equation}
    \mathcal{L}_i = \sum_{(x,y) \in \mathcal{D}_i} \left[ \mathcal{L}_{\text{pre-reg}} + \beta \sum_{k=1}^{K} \|\phi_i^{(k)}\|_2^2 + \gamma H(\alpha_i(x)) \right]
    \label{eq:reg_loss}
\end{equation}

where
\begin{equation}
    H(\alpha_i(x)) = -\sum_{k=1}^{K} \alpha_i^{(k)}(x) \log \alpha_i^{(k)}(x).
\end{equation}

$H(\alpha_i(x))$ is the entropy regularization term and $\beta$ and $\gamma$ are hyperparameters controlling regularization strength.

\subsubsection{Diffusion-based Input Purification}

The purification module mitigates adversarial perturbations by emulating a forward noising process that progressively corrupts the input, followed by a learned reverse denoising process that reconstructs the underlying clean image. This mechanism is motivated by the manifold hypothesis, which asserts that high-dimensional data, such as natural images, predominantly reside on a low-dimensional manifold. Adversarial noise typically displaces an image away from this manifold into a low-density region. By applying the forward diffusion process, the structured adversarial components are degraded, and the subsequent reverse process, trained to map corrupted inputs back onto the learned data manifold, restores a clean representation of the image~\cite{pmlr-v162-nie22a}.

This module employs a U-Net architecture optimized for memory efficiency:
\begin{itemize}
    \item \textbf{Encoder:} 4 downsampling blocks with DoubleConv layers.
    \item \textbf{Decoder:} 4 upsampling blocks with skip connections.
    \item \textbf{Hidden Channels:} 256 channels.
    \item \textbf{Time Embedding:} Multi-layer perceptron with SiLU activation, responsible for injecting temporal information into the network.
\end{itemize}

To integrate temporal information, the model employs a time embedding mechanism that maps discrete timesteps into a continuous latent representation. This is achieved by first applying sinusoidal positional encoding, followed by a linear transformation and nonlinear activation:

\begin{equation}
\text{TimeEmbed}(t) = \text{MLP}(\text{SiLU}(\text{Linear}(\text{SinCos}(t)))),
\end{equation}

where $\text{SinCos}(t)$ denotes the sinusoidal embedding of timestep $t$:

\begin{equation}
\text{SinCos}(t)_i = \begin{cases}
\sin(t / 10000^{2i/d}) & \text{if } i \text{ is even} \\
\cos(t / 10000^{(2i-1)/d}) & \text{if } i \text{ is odd}
\end{cases}
\end{equation}

The forward process progressively corrupts the input with Gaussian noise according to:
\begin{equation}
    q(x_t | x_{t-1}) = \mathcal{N}(x_t; \sqrt{1-\beta_t}x_{t-1}, \beta_t I),
\end{equation}
with a linear variance schedule $\{\beta_t\}_{t=1}^T$ where $\beta_1 = 10^{-4}$, $\beta_T = 2 \times 10^{-2}$, and $T = 1000$ timesteps.


The reverse denoising process is modeled as:
\begin{equation}
    p_\phi(x_{t-1} | x_t) = \mathcal{N}(x_{t-1}; \mu_\phi(x_t, t), \Sigma_\phi(x_t, t)),
\end{equation}
where
\begin{equation}
    \mu_\phi(x_t, t) = \frac{1}{\sqrt{\alpha_t}}\left(x_t - \frac{\beta_t}{\sqrt{1-\bar{\alpha}_t}} \epsilon_\phi(x_t, t)\right),
\end{equation}
$\alpha_t = 1 - \beta_t$, and $\bar{\alpha}_t = \prod_{s=1}^{t} \alpha_s$.

The model is trained using the denoising score matching loss:
\begin{equation}
    \mathcal{L}_{\mathrm{diff}} =
    \mathbb{E}_{x_0, \epsilon \sim \mathcal{N}(0,I), \, t \sim U(1,T)}
    \big[\|\epsilon - \epsilon_\phi(x_t, t)\|^2\big],
\label{eq:diff_loss}
\end{equation}
where 
\begin{equation}
    x_t = \sqrt{\bar{\alpha}_t}x_0 + \sqrt{1-\bar{\alpha}_t}\epsilon.
\end{equation}

\subsubsection{Masked Autoencoder Detection}

To detect adversarial perturbations, we employ a MAE built on a ViT backbone. The encoder is composed of 12 transformer layers with an embedding dimension of 512 and 8 self-attention heads per layer. The decoder consists of 8 transformer layers with the same embedding dimension and attention configuration. For CIFAR-10 dataset which has small spatial resolution, we use a patch size of \(4\times4\), while for the higher-resolution Br35H dataset, a \(16\times16\) patch size is adopted.

An input image $x \in \mathbb{R}^{H \times W \times C}$ is first divided into non-overlapping patches of size $p \times p$, which are then flattened and projected into a $D$-dimensional embedding space:
\begin{equation}
\text{PE}: \mathbb{R}^{H \times W \times C} \rightarrow \mathbb{R}^{N \times D},
\end{equation}
where 
$$
P_{i,j} = \text{Flatten}(x[i \cdot p:(i+1) \cdot p,\, j \cdot p:(j+1) \cdot p, :]),
$$
\begin{equation}
\quad N = \frac{HW}{p^2}.
\end{equation}

During training, a high masking ratio is applied to encourage robust feature learning. Specifically, a random binary mask \(\mathcal{M}\) is sampled according to:
\begin{equation}
\mathcal{M} \sim \text{Bernoulli}(1-r)^{P}, \quad r = 0.75,
\end{equation}
where \(P\) is the total number of image patches and \(r\) is the masking ratio. Only the unmasked patches are visible to the encoder, with the decoder reconstructing the masked regions.

The MAE is optimized using a mean squared error (MSE) loss computed solely on the masked patches:
\begin{equation}
\mathcal{L}_{\mathrm{MAE}} = \frac{1}{|\mathcal{M}^c|} \sum_{p \in \mathcal{M}^c} \| x_p - \hat{x}_p \|^2,
\label{eq:mae_loss}
\end{equation}
where \(\mathcal{M}^c\) denotes the set of masked patches, \(x_p\) is the ground truth pixel value for patch \(p\), and \(\hat{x}_p\) is its reconstruction.

Once trained, the MAE serves as a detection module by measuring the average reconstruction error across all patches:
\begin{equation}
s_{\mathrm{det}}(x) = \frac{1}{P} \sum_{p=1}^{P} \| x_p - \hat{x}_p \|^2.
\end{equation}

Rather than adopting a fixed threshold for all datasets, we employ a rank-based adaptive criterion. 
For each dataset, the reconstruction scores $\{s_{\mathrm{det}}(x)\}$ are sorted, and the top-$\kappa\%$ of samples 
with the largest reconstruction errors are flagged as potentially adversarial and forwarded to the diffusion-based purification stage. 
The percentile $\kappa$ is selected according to the empirical distribution of reconstruction errors to capture the heavy-tail region 
that typically corresponds to abnormal or out-of-distribution inputs. 
This strategy avoids manual calibration of a universal threshold $\tau$ and ensures that detection sensitivity remains stable 
across datasets with different dimensionalities and noise characteristics. 
In our experiments, $\kappa=5\%$ for CIFAR-10 and $\kappa=18\%$ for Br35H.  

The underlying rationale is that the MAE, trained solely on benign samples, captures a compact representation of the natural image manifold. 
When presented with adversarial inputs, which usually lie outside this manifold, the reconstruction process operates on latent representations 
that are out-of-distribution, producing significantly higher reconstruction errors. 
Consequently, the reconstruction score $s_{\mathrm{det}}(x)$ serves as a reliable measure of deviation from the learned data manifold, 
and the rank-based selection offers a data-driven and interpretable detection mechanism.

\subsection{Integrated Defense Pipeline}
\label{sec:integrated_defense}

At inference time, our framework integrates adversarial detection, conditional purification, and personalized prediction into a unified pipeline. An incoming sample is first evaluated by the MAE-based detector to produce a confidence score $s_{\text{det}}(x)$. If the score exceeds a threshold $\tau$, the sample undergoes diffusion-based purification; otherwise, it is passed directly to the predictor. The purification depth is adaptively determined from $s_{\text{det}}(x)$, enabling stronger denoising for heavily perturbed inputs while minimizing distortion to clean samples. Finally, the purified (or unaltered) input is classified using the Personalized Federated Defense model, ensuring robustness without sacrificing diagnostic fidelity. This tight coupling of detection, adaptive purification, and personalized learning allows the system to respond proportionally to varying attack intensities, balancing robustness and accuracy in federated medical imaging scenarios.

\subsection{Federated Training Algorithm}

Our training pipeline consists of three coordinated phases: diffusion model pretraining, adversarial detection training, and personalized federated model optimization. This design ensures that each defense component is independently effective while remaining seamlessly integrated into the overall framework.

\subsubsection{Local Client Updates}

Each client \( i \) maintains three local experts and corresponding attention networks. Local training proceeds for \( E = 15 \) epochs per communication round. For each mini-batch \( \mathcal{B}_i \) from the local dataset \( \mathcal{D}_i \), the client:
\begin{enumerate}
    \item Updates the parameters of each expert network via stochastic gradient descent on the classification loss.
    \item Updates the attention network parameters to refine the input-dependent mixture weights.
\end{enumerate}
This procedure allows clients to adapt their experts to both benign and adversarially perturbed inputs.

\subsubsection{Global Aggregation}

Following local updates, the server aggregates model parameters using a sample-size-weighted averaging scheme (FedAvg). For each expert \( k \), the global parameters are computed as:
\begin{equation}
\theta_{\mathrm{global}}^{(k)} = \sum_{i=1}^{N} \frac{n_i}{n} \theta_i^{(k)},
\end{equation}
where \( n_i \) is the number of training samples at client \( i \), and \( n = \sum_{i=1}^N n_i \).

\subsubsection{Training Phases and Integration}

\paragraph{Phase 1: Diffusion Model Pretraining}  
For each dataset in the experimental suite, we pretrain a memory-efficient U-Net based diffusion model over 50 epochs. As described in Algorithm~\ref{alg:phase1}, this stage employs the denoising score matching objective (Eq.~\ref{eq:diff_loss}), yielding a dataset-specific purification model. The resulting model is able to suppress adversarial perturbations while maintaining essential image details.

\begin{algorithm}[h!]
\caption{Phase 1: Diffusion Model Pretraining}
\label{alg:phase1}
\begin{algorithmic}[1]
\State \textbf{Input:} Datasets $\mathcal{D}_{d}$ for $d \in \{\text{CIFAR-10, \dots, Br35H}\}$.
\State \textbf{Output:} Trained purification models $\{\epsilon_{\phi_d}\}$.
\For{each dataset $d$}
    \State Initialize diffusion model parameters $\phi_d$.
    \For{epoch = 1 to 50}
        \State Sample a mini-batch of clean images $x_0$ from $\mathcal{D}_{d}$.
        \State Update $\phi_d$ by descending the gradient of $\mathcal{L}_{\mathrm{diff}}$ (from Eq.~\ref{eq:diff_loss}).
    \EndFor
\EndFor
\end{algorithmic}
\end{algorithm}

\paragraph{Phase 2: MAE Detector Training}  
In the second stage (Algorithm~\ref{alg:phase2}), we initialize the MAE with an embedding dimension of \(512\) and train it for 30 epochs using the reconstruction-based detection loss (Eq.~\ref{eq:mae_loss}). This detector produces adversarial likelihood scores that are subsequently integrated into the adaptive purification process.

\begin{algorithm}[h!]
\caption{Phase 2: MAE Detector Training}
\label{alg:phase2}
\begin{algorithmic}[1]
\State \textbf{Input:} Datasets $\mathcal{D}_{d}$ for $d \in \{\text{CIFAR-10, \dots, Br35H}\}$.
\State \textbf{Output:} Trained detector models $\{\text{MAE}_{\psi_d}\}$.
\For{each dataset $d$}
    \State Initialize MAE detector parameters $\psi_d$.
    \For{epoch = 1 to 30}
        \State Sample a mini-batch of clean images $x$ from $\mathcal{D}_{d}$.
        \State Update $\psi_d$ by descending the gradient of $\mathcal{L}_{\mathrm{MAE}}$ (from Eq.~\ref{eq:mae_loss}).
    \EndFor
\EndFor
\end{algorithmic}
\end{algorithm}

\paragraph{Phase 3: Federated Model Optimization}  
The final stage involves federated training of the personalized classifier, as outlined in Algorithm~\ref{alg:phase3}. Each federated round (\(T=20\)) begins with clients downloading the current global model. Local training is then performed using the attention regularized cross-entropy loss on the final aggregated prediction (Eq.~\ref{eq:reg_loss}), with an optional defense step: whenever a sample's detection score exceeds the threshold, it is purified before entering the classifier. After completing local updates, client parameters are uploaded to the server, which applies FedAvg aggregation. Performance is monitored at every round through evaluation on both clean and PGD-perturbed test sets.

\begin{algorithm}[h!]
\caption{Phase 3: Personalized Federated Learning}
\label{alg:phase3}
\begin{algorithmic}[1]
\State \textbf{Input:} Number of clients $N$, communication rounds $T$, local epochs $E$.
\State \textbf{Output:} Trained global models $\{\theta_{\text{global}}^{(k)}\}_{k=1}^K$.
\State Initialize global expert models $\theta_{\text{global}}^{(k)}$ for $k=1, \dots, K$.
\For{round $t = 1$ to $T$}
    \For{each client $i = 1$ to $N$ \textbf{in parallel}}
        \State $\theta_i^{(k)} \leftarrow \theta_{\text{global}}^{(k)}$ for $k=1, \dots, K$ 
        \For{local epoch $e = 1$ to $E$}
            \State Sample mini-batch $\mathcal{B}_i = \{(x_j, y_j)\}_{j=1}^{B}$ from $\mathcal{D}_i$.
            \State \textit{// Optional: Apply defense during local training}
            \For{each sample $(x_j, y_j) \in \mathcal{B}_i$}
                \If{$s_{\mathrm{det}}(x_j) > \tau$}
                    \State $x_j \leftarrow \mathrm{DiffusionPurify}(x_j)$
                \EndIf
            \EndFor
            \State Update local models $\{\theta_i^{(k)}\}_{k=1}^K$ by descending the gradient of $\mathcal{L}_{\text{client}}$.
        \EndFor
        \State Upload $\{\theta_i^{(k)}\}_{k=1}^K$ to the server.
    \EndFor
    \State \textbf{Server Aggregation:}
    \For{$k=1$ to $K$}
        \State $\theta_{\text{global}}^{(k)} \leftarrow \sum_{i=1}^{N} \frac{n_i}{n} \theta_i^{(k)}$ \Comment{FedAvg}
    \EndFor
\EndFor
\end{algorithmic}
\end{algorithm}

\section{Experiments and Results}\label{sec:experiments}
\subsection{Datasets}
\label{sec:datasets}
We evaluate the proposed framework on two datasets, combining standard computer vision benchmarks with a specialized medical imaging collection. This selection facilitates assessment of both generalizability and domain-specific performance.

For baseline comparison, we employ \textbf{CIFAR-10}~\cite{krizhevsky2009learning}, comprising 60{,}000 color images of size $32\times32$ across 10 classes.  

Our primary focus lies in the medical domain, represented by the \textbf{Br35H} (Brain Tumor Detection 2020) dataset~\cite{hamada2020br35h}, containing 3{,}000 brain MRI scans annotated for binary classification (tumor vs. healthy). All Br35H images were resized to $224\times224$ pixels and normalized to the $[0,1]$ range. To reduce overfitting and improve robustness, we applied standard data augmentation, including random rotations, flips, and brightness adjustments, during training.

\subsection{Implementation Details}
All experiments were implemented using PyTorch~2.0+ with CUDA~11.8+ on an NVIDIA RTX 3090 GPU with 24GB VRAM. We employed MoblieNetV2~\cite{sandler2018mobilenetv2} and ResNet-18~\cite{he2016deep} for $10$ federated clients. 

\subsection{Evaluation Metrics}
The evaluation framework covers robustness, detection, and efficiency metrics.  
Clean accuracy and adversarial accuracy are defined as:  
\begin{subequations}\label{eq:acc_defs}
\begin{align}
\mathrm{Acc}_{\mathrm{clean}}
 &= \frac{1}{|\mathcal{T}|}\sum_{(x,y)\in\mathcal{T}}\mathbf{1}[f(x)=y], \label{eq:acc_clean}\\
\mathrm{Acc}_{\mathrm{adv}}
 &= \frac{1}{|\mathcal{T}|}\sum_{(x,y)\in\mathcal{T}}\mathbf{1}[f(x_{\text{adv}})=y]. \label{eq:acc_adv}
\end{align}
\end{subequations}

\subsection{Results and Baseline Comparison}
\begin{figure}[!t]
    \centering
    \subfloat[BR35H]{%
        \includegraphics[width=\linewidth]{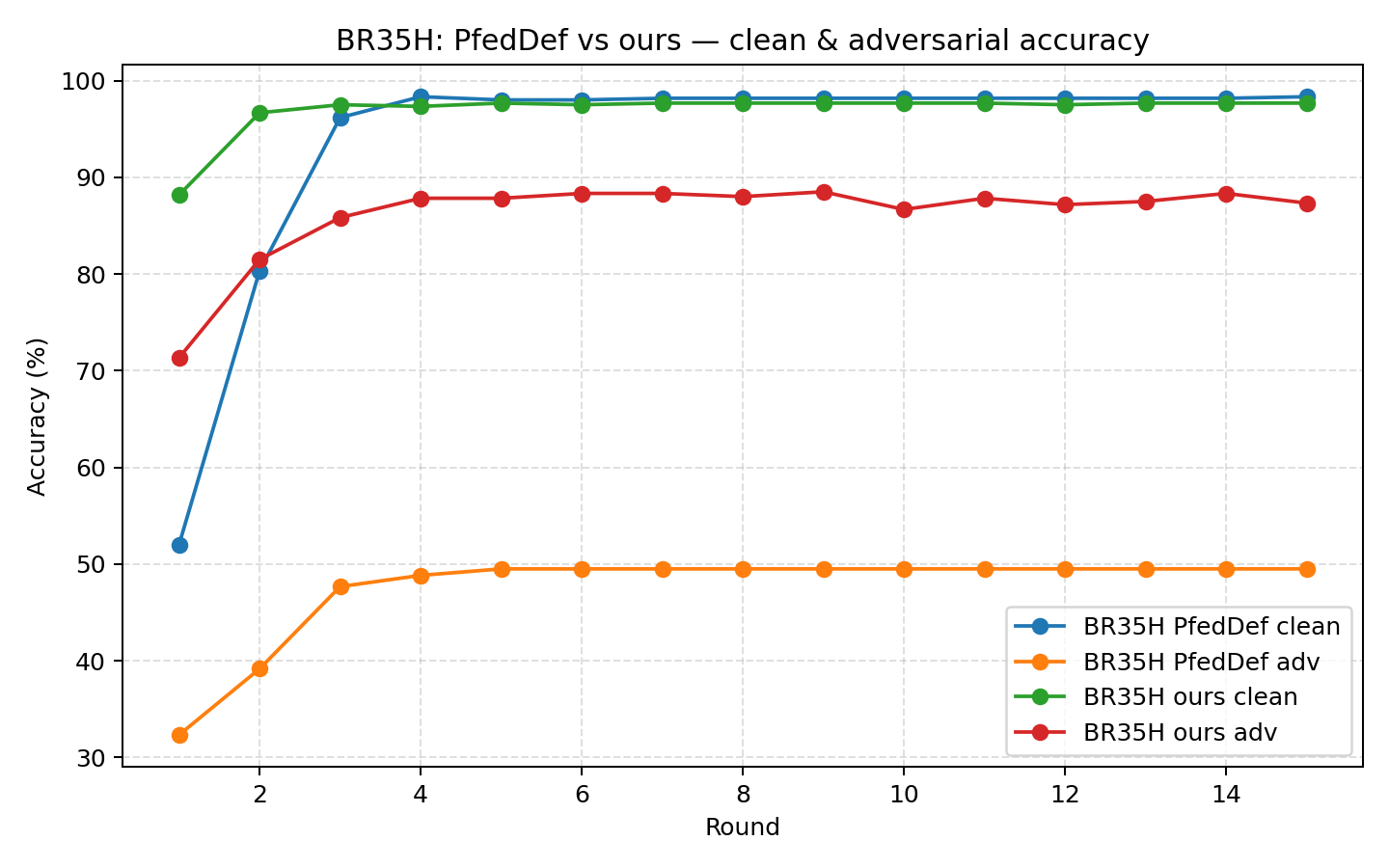}%
        \label{fig:conv_br35h}}
    \par\vspace{4pt}
    \subfloat[CIFAR-10]{%
        \includegraphics[width=\linewidth]{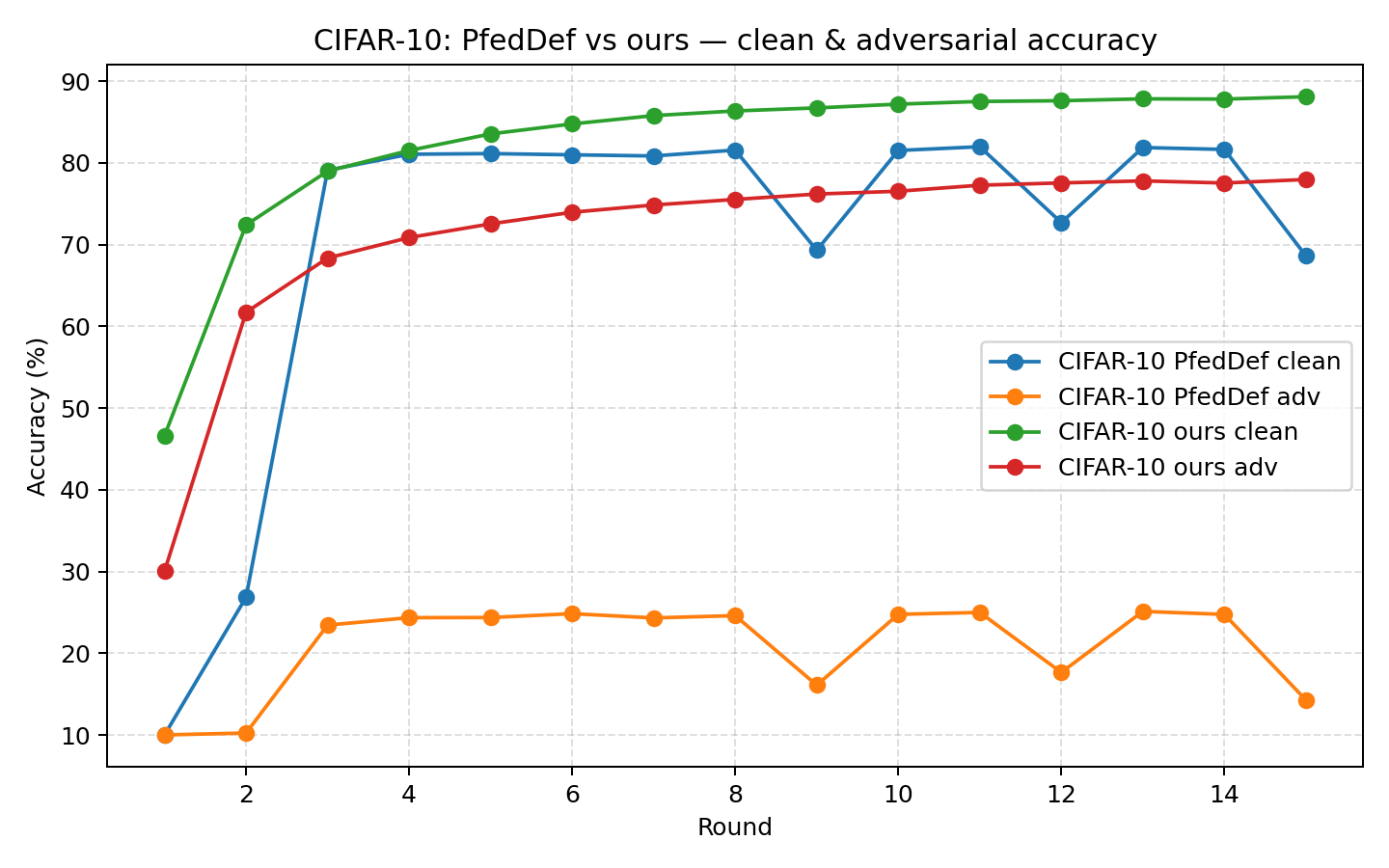}%
        \label{fig:conv_cifar10}}
    \caption{Convergence of clean and adversarial accuracy across federated rounds for the PfedDef vs.\ our method. (a) BR35H; (b) CIFAR-10.}
    \label{fig:convergence}
\end{figure}

To evaluate our framework, we compare it against a comprehensive set of state-of-the-art \textbf{test-time} adversarial defenses, as our primary contribution is an efficient, adaptive defense that operates exclusively at inference. we include two key federated baselines. The first is FedAvg, which provides a standard benchmark for federated learning without adversarial robustness. The second is pFedDef~\cite{kim2023characterizing}, a personalized federated defense framework that combines adversarial training with FL. pFedDef serves as our primary comparable baseline in the federated setting, offering a direct point of reference for evaluating robustness-oriented methods. The comparison between our method and pFedDef is presented in Fig.~\ref{fig:convergence}.

As pFedDef represents the most directly comparable federated defense, we further provide a dedicated comparison against it. Specifically, we evaluate pFedDef under different PGD attack step sizes ($\alpha = 0.007, 0.01$) on CIFAR-10, and also report results on Br35H, which is the main dataset of our study. Both clean and adversarial accuracies are reported for the datasets in Table~\ref{tab:paper_cifar_comparison} and \ref{tab:pfeddef_br35h}.

While pFedDef achieves moderate improvements in robustness compared to standard FL methods, our results highlight significant limitations in both scalability and defense effectiveness. On CIFAR-10 (Table~\ref{tab:paper_cifar_comparison}), we replicated the exact configuration used in the original pFedDef paper, including MobileNetV2 as the client model and $L_2$-norm PGD attacks. Under these conditions, pFedDef achieves a clean accuracy of $66.0\%$ and adversarial accuracy of $48.0\%$. In contrast, MedFedPure maintains substantially higher clean performance ($82.31\%$) while offering stronger robustness, reaching an adversarial accuracy of $69.13\%$. This indicates that our design effectively mitigates the trade-off between clean accuracy and robustness that is typically observed in adversarially trained federated models.

In Table~\ref{tab:pfeddef_br35h}, we extend the evaluation to a stronger setting, employing $L_\infty$-norm PGD attacks and using ResNet-18 as the backbone. On CIFAR-10, MedFedPure continues to demonstrate large gains in adversarial robustness compared to pFedDef. The results on Br35H are even more telling: while pFedDef attains a clean accuracy of $98.33\%$ and an adversarial accuracy of $49.50\%$, MedFedPure maintains nearly identical clean performance ($97.67\%$) while achieving a substantial boost in robustness ($87.33\%$). This large margin is particularly significant given that Br35H is the primary dataset of our study, directly underscoring the practical value of our framework for robust medical image analysis in federated environments.

Taken together, these results show that MedFedPure consistently outperforms both centralized test-time defenses and federated adversarial training approaches. Importantly, the improvements on Br35H, a clinically relevant medical imaging dataset, highlight the ability of our framework to maintain both accuracy and robustness in sensitive diagnostic tasks. This suggests that the proposed defense is not only effective in standard benchmarks such as CIFAR-10 but also directly transferable to medical image analysis settings, where resilience against perturbations is critical. 

\begin{table}[!htbp]
\centering
\caption{Performance of models under different federated adversarial training algorithms on CIFAR-10 against internal  $L_2$-norm PGD attacks. \newline
The attack was run with step size $\alpha = 0.01$, number of steps $K = 10$, and perturbation budget $\epsilon = 0.1$. Clients used MobileNetV2.}
\label{tab:paper_cifar_comparison}
\begin{tabular}{lcc}
\toprule
\textbf{Method} & \textbf{Clean Acc. (\%)} & \textbf{Adv. Acc. (\%)} \\
\midrule
\multicolumn{3}{l}{\textit{Models without Adversarial Training}} \\
\midrule
FedAvg       & $93.0$ & $0.0$ \\
Local Training       & $38.0$ & $33.0$ \\
FedEM\cite{marfoq2021federated}       & $76.0$ & $13.0$ \\
\midrule
\multicolumn{3}{l}{\textit{Models with Adversarial Training}} \\
\midrule
FAT (FedAvg with AT)         & $80.0$ & $30.0$ \\
Local Adv.   & $30.0$ & $28.0$ \\
pFedDef      & $66.0$ & $48.0$ \\
\midrule
\textbf{MedFedPure} & \textbf{82.31} & \textbf{51.43}\\
\bottomrule
\end{tabular}
\end{table}

\begin{table}[!htbp]
\centering
\caption{Comparison of pFedDef and MedFedPure on Br35H and CIFAR-10 under $L_\infty$-norm PGD. Our approach demonstrates stronger adversarial robustness while preserving clean accuracy.\newline
The attack was run with number of steps $K = 7$ and perturbation budget $\epsilon = 0.015$. Clients used ResNet-18.}
\label{tab:pfeddef_br35h}
\begin{tabular}{lp{0.6cm}ccc}
\toprule
\textbf{Dataset} & \textbf{Step Size ($\alpha$)} & \textbf{Method} & \textbf{Clean Acc. (\%)} & \textbf{Adv. Acc. (\%)} \\
\midrule
\multirow{2}{*}{CIFAR-10} 
    & \multirow{2}{*}{0.007} & pFedDef              & $68.64$ & $14.28$ \\
    &                        & \textbf{MedFedPure}  & \textbf{88.11} & \textbf{77.98} \\
\midrule
\multirow{2}{*}{Br35H} 
    & \multirow{2}{*}{0.003} & pFedDef              & $98.33$ & $49.50$ \\
    &                        & \textbf{MedFedPure}  & \textbf{97.67} & \textbf{87.33} \\
\bottomrule
\end{tabular}
\end{table}

\section{Conclusion}\label{sec:conclusion}
This paper addressed the vulnerability of federated learning-based medical imaging models to inference-time adversarial attacks. We introduced MedFedPure, a client-side defense framework combining a personalized federated classifier, an MAE-based detector, and an adaptive diffusion purification module.
On the Br35H brain MRI dataset, MedFedPure improved adversarial robustness from 49.50\% to 87.33\% while maintaining a clean accuracy of 97.67\%, outperforming existing federated defenses. These results confirm that the framework can preserve diagnostic reliability even under strong adversarial interference.

From a medical perspective, such resilience is essential: MRI-based tumor detection requires not only high baseline accuracy but also protection against subtle perturbations that could cause misdiagnosis or treatment delays. Moreover, by operating entirely on the client side, MedFedPure ensures privacy-preserving collaboration across institutions without sharing raw patient data, addressing key regulatory and ethical constraints. Overall, MedFedPure offers a robust, accurate, and privacy-compatible solution for secure deployment of federated AI systems in brain tumor detection workflows.

\section{Future Works}
While MedFedPure demonstrates promising robustness and clinical relevance, several directions remain open for exploration. 

First, future research should extend evaluation to larger, more diverse, and multi-modal medical datasets (e.g., CT, PET, histopathology, and multi-sequence MRI). Such validation across heterogeneous sources will help assess the framework's generalizability and ensure its clinical applicability beyond Br35H. 

Second, our current work primarily targets $\ell_\infty$-bounded adversarial perturbations. Broader threat modeling is needed to encompass adaptive attacks, backdoor triggers, model inversion, and membership inference in federated environments. Designing defenses that can withstand coordinated and adaptive adversaries remains an open challenge.

Third, improving the efficiency of the diffusion purification stage is essential for deployment in real-time medical workflows. Future work may explore lightweight generative alternatives (e.g., diffusion distillation, score-based denoisers, or generative adversarial purification) to reduce latency while preserving purification strength. Integration with hardware-aware optimization could further adapt MedFedPure to resource-constrained hospital devices.

Fourth, personalization remains an evolving frontier. Our mixture-of-experts design offers adaptability, but additional strategies such as meta-learning, federated clustering, or client-level trust modeling could be integrated to handle extreme heterogeneity in client data distributions. Exploring the synergy between personalization and security may yield new defense paradigms tailored to real-world federated hospital networks.

Fifth, explainability and interpretability should be emphasized in future extensions. Incorporating explainable AI modules that can visualize or rationalize detection and purification decisions would improve clinical trust and facilitate adoption by radiologists and medical practitioners.

Finally, translating MedFedPure from experimental validation into a deployable system requires collaboration with healthcare institutions. This involves testing under federated networks with realistic constraints (irregular participation, variable bandwidth, heterogeneous hardware) and assessing usability in live diagnostic settings. Such translational studies will be critical for bridging the gap between algorithmic advances and safe, trustworthy clinical AI deployment.

In summary, future research can build upon MedFedPure by (i) validating across broader datasets and modalities, (ii) defending against richer adversarial scenarios, (iii) reducing latency with efficient purification, (iv) advancing personalization strategies, (v) adding interpretability, and (vi) pursuing real-world clinical deployment. Addressing these directions will further strengthen the robustness, scalability, and trustworthiness of federated medical AI.

\printbibliography
\end{document}